\documentclass[10pt,twocolumn,letterpaper]{article}

\usepackage{iccv}
\usepackage{times}
\usepackage{epsfig}
\usepackage{graphicx}
\usepackage{amsmath}
\usepackage{amssymb}

\usepackage{booktabs}
\usepackage[ruled,linesnumbered]{algorithm2e}
\usepackage{bm}
\usepackage{makecell}
\usepackage{diagbox}
\usepackage{multirow}
\usepackage[pagebackref=true,breaklinks=true,letterpaper=true,colorlinks,bookmarks=false]{hyperref}
\usepackage[dvipsnames]{xcolor}
\usepackage{setspace}
\usepackage[accsupp]{axessibility}

\newcommand{\myref}[1]{Eq.\ref{#1}}
\newtheorem{theorem}{Theorem}
\usepackage[capitalize]{cleveref}
\crefname{section}{Sec.}{Secs.}
\Crefname{section}{Section}{Sections}
\Crefname{table}{Table}{Tables}
\crefname{table}{Tab.}{Tabs.}

 \iccvfinalcopy 

\def\iccvPaperID{4122} 

\ificcvfinal\fi

\begin{document}


\title{ Semi-Supervised Learning via Weight-aware Distillation under Class Distribution Mismatch}

\author{Pan Du\textsuperscript{1,2}, Suyun Zhao\textsuperscript{1,2,*}, Zisen Sheng\textsuperscript{1, 2}, Cuiping Li\textsuperscript{1,2}, Hong Chen\textsuperscript{1,2}\\
	Key Lab of Data Engineering and Knowledge Engineering of MOE Renmin University of China\textsuperscript{1}\\ Renmin University of China, Beijing, China\textsuperscript{2}\\
	{\tt\small \{du\_pan,shengzisen\}@163.com,  \tt\small\{zhaosuyun, zisen, licuiping, chong\}@ruc.edu.cn
}
}

\maketitle
\ificcvfinal\thispagestyle{empty}\fi
\footnotetext{*Corresponding Author}

	\begin{abstract}\label{sec:abs}

Semi-Supervised Learning (SSL) under class distribution mismatch aims to tackle a challenging problem wherein unlabeled data contain lots of unknown categories unseen in the labeled ones. 
In such mismatch scenarios, traditional SSL suffers severe performance damage due to the harmful invasion of the instances with unknown categories into the target classifier. 
In this study, by strict mathematical reasoning, we reveal that the SSL error under class distribution mismatch is composed of pseudo-labeling error and invasion error, both of which jointly bound the SSL population risk. To alleviate the SSL error, we propose a robust SSL framework called Weight-Aware Distillation (WAD) that, by weights, selectively transfers knowledge beneficial to the target task from unsupervised contrastive representation to the target classifier. Specifically, WAD captures adaptive weights and high-quality pseudo-labels to target instances by exploring point mutual information (PMI) in representation space to maximize the role of unlabeled data and filter unknown categories. Theoretically, we prove that WAD has a tight upper bound of population risk under class distribution mismatch. Experimentally, extensive results demonstrate that WAD outperforms five state-of-the-art SSL approaches
and one standard baseline on two benchmark datasets, CIFAR10 and CIFAR100, and an artificial cross-dataset. 
 The code is
available at \url{https://github.com/RUC-DWBI-ML/research/tree/main/WAD-master}.

\end{abstract}

\section{Introduction}\label{sec:intro}

\begin{figure}[t]
	\centering
	\includegraphics[width=1\linewidth]{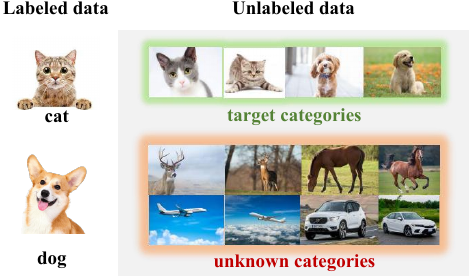}
	\caption{Example of class distribution mismatch. The unlabeled data contains categories that are unseen in labeled ones. }
	\label{fig:class_distribution}
	\vspace{-0.4cm}
\end{figure}

Deep neural networks (DNNs) have achieved remarkable success in fully-supervised learning tasks. However, sufficient labeled data are usually unavailable in real applications due to the expensive annotation cost or even domain-specific knowledge required ~\cite{ssl_cao2021open, 28_SSL_USAD, Du_2022_tpami, Du_2021_ICCV}. Semi-supervised learning (SSL), as a powerful weakly-supervised technique, provides an effective way to improve DNNs by exploiting massive unlabeled data, and then it weakens the demand for human annotation~\cite{chapelle2009semi, grandvalet2004semi, laine2016temporal, tarvainen2017mean}. 
Generally, traditional SSL approaches assume that the labeled and unlabeled instances share the same class distribution, i.e., they come from identical categories. However, in real scenarios, this assumption hardly holds as unlabeled data inevitably contains lots of categories unseen in labeled ones. For instance, if unlabeled data are collected from the internet using keywords ``cat'' and ``dog'' (target categories), they may contain instances unrelated to these categories, such as ``deer,'' ``horse,'' or ``airplane''(unknown categories), as shown in Figure~\ref{fig:class_distribution}. Similar scenarios occur in medical diagnoses~\cite{28_SSL_USAD, 13_SSL_DS3L} and house annotations of remote-sensing images~\cite{Du_2022_tpami,Du_2021_ICCV}. SSL in such mismatch scenarios is called SSL under class distribution mismatch ~\cite{Du_2022_tpami,13_SSL_DS3L}. 

\begin{figure}[t]
	\centering
	\includegraphics[width=1\linewidth]{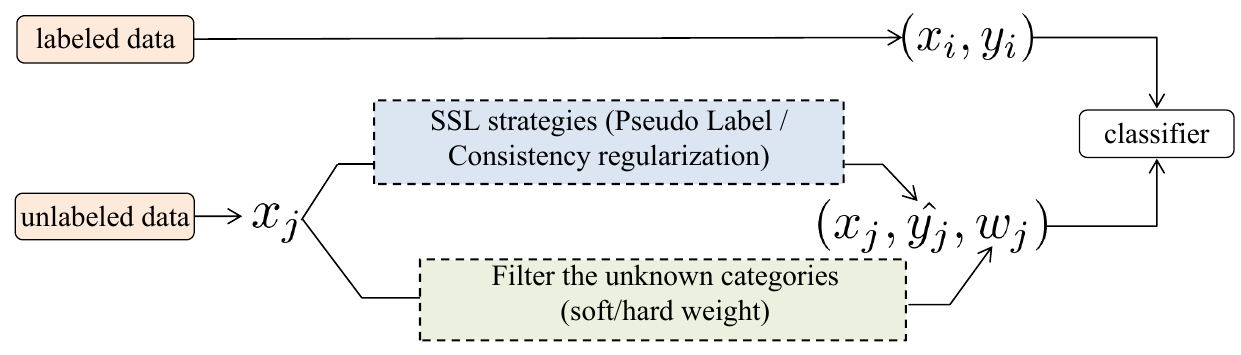}
	\caption{The paradigm of SSL under class distribution mismatch. }
	\label{fig:ssl_pra}
\end{figure}

Under class distribution mismatch, some SSL approaches~\cite{ssl_cao2021open, 28_SSL_USAD, 13_SSL_DS3L, ssl_huang2021trash, yang2022class} have been proposed. Usually, most of them exploit pseudo-labeling or consistency regularization to expand \textcolor{black}{the} labeled pool, as well as filter instances with unknown categories by weights, just as shown in Figure~\ref{fig:ssl_pra}. 
UASD~\cite{28_SSL_USAD} and T2T~\cite{ssl_huang2021trash} filter out \textcolor{black}{the instances with} unknown categories by leveraging a hard weight, i.e., a threshold, on the accumulated network's output or the out-of-distribution score.
Although these two approaches reduce the invasion of unknown categories, it is inevitable to keep off amounts of unlabeled instances with target categories. 
\textcolor{black}{Instead of hard weights,} Guo et al.~\cite{13_SSL_DS3L} assign a soft weight to the unlabeled instances according to the consistent empirical risk loss. In such case, many instances with unknown categories tend to have consistent outputs and get high weights, just \textcolor{black}{as shown in Appendix 4.3,} and then they may invade the target classifier and impair its performance. 

Moreover, the existing SSL approaches with consistency regularization and pseudo-labeling heavily rely on the \textcolor{black}{performance of the target} classifier.
 Both ~\cite{13_SSL_DS3L} and~\cite{ssl_huang2021trash} annotate pseudo labels by leveraging the prediction of the target classifier in training. Once the target classifier trained on limited labeled instances is biased by some instances with unknown categories, the subsequently updated target classifier may allow more unknown instances 
to invade. 
Accordingly, it is promising to propose a novel SSL approach that captures pseudo labels from representations produced by all available data rather than an immature classifier. 


In this study, by strict theoretical analyses, we decouple the SSL error under class distribution mismatch into pseudo-labeling error and invasion error (\textcolor{black}{seen in Subsection~\ref{sbusec:population}}).
According to this discovery, a robust SSL framework called weight-aware distillation (WAD) is then proposed to distill pseudo labels and weights from the representation space to the target classifier. Unlike the conventional distillation approaches~\cite{bucilua2006model, hinton2015distilling, nayak2019zero} that simply train the student model using the prediction probability of the teacher model, WAD is a weight-aware distillation framework that adapts to mismatch problems.
 Specifically, we learn the representations from labeled and unlabeled data by unsupervised contrastive coding, as the teacher model. Then WAD captures  adaptive weights as well as high-quality pseudo labels from the teacher model by leveraging point mutual information(PMI), 
 and thus, the target classifier could selectively utilize the instances from target categories while filtering the ones with unknown categories.

Our main contributions are listed as follows.
\begin{itemize}
	\item[\romannumeral1 )]
	We theoretically analyze the population risk in an SSL manner and reveal that the SSL error under class distribution mismatch is jointly controlled by pseudo-labeling error and invasion error.
	\item[\romannumeral2 )] 
	 We propose a distillation-based SSL framework, WAD, that captures weights as well as pseudo labels  from robust representations to the target classifier to filter unknown categories and make full use of targeted unlabeled instances as well. 
	\item[\romannumeral3 )]
	Theoretically, we verify that the population risk of WAD is tightly bounded. Experimentally, WAD outperforms five state-of-the-art SSL approaches and one standard baseline on several datasets. 
\end{itemize}

\section{Related Work}\label{sec:related}

\begin{figure*}[ht]
	\centering
	\includegraphics[width=1\linewidth]{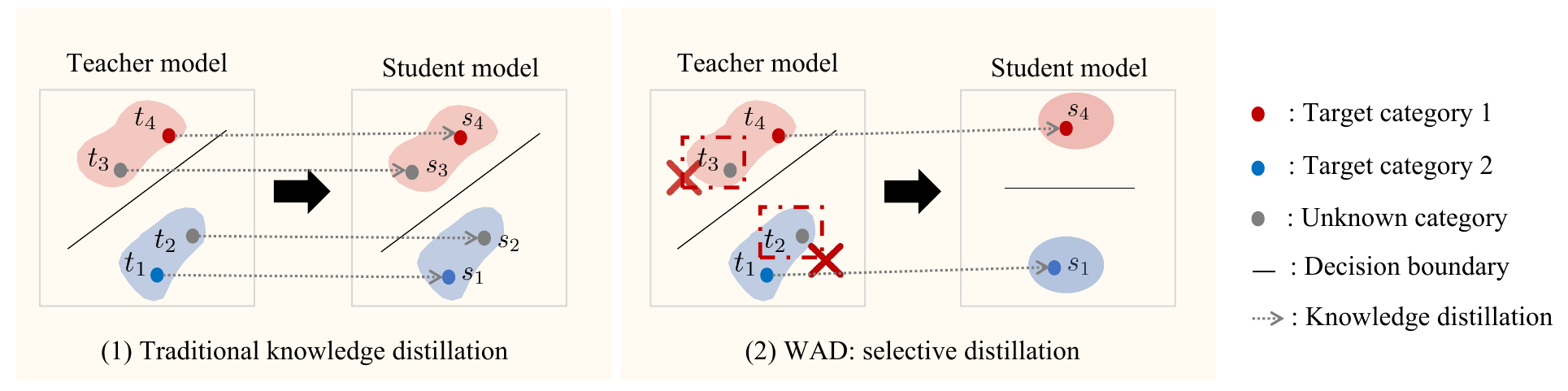}
	\caption{The selective distillation in WAD. Traditional distillation methods force the output of the student model ($s_i$) to align with that of the teacher model ($t_i$), resulting in a wrong decision boundary. However, WAD selectively distills benefit knowledge to student and filter the negatives, such as $s_2$ and $s_3$, by weights to rectify the decision boundary and solve the problem of class distribution mismatch.}
	\label{fig:x}
\end{figure*}

This section reviews the SSL \textcolor{black}{approaches} under class distribution match and mismatch. 
For contrastive learning, \textcolor{black}{please refer to Appendix 1.}

\noindent {\bf Semi-Supervised Learning.} 
The traditional SSL strategies include entropy minimization, consistency regularization, and pseudo-label. Entropy minimization~\cite{grandvalet2004semi} incorporates unlabeled data in supervised learning by minimizing the entropy of the unlabeled instance's prediction. The consistency regularization~\cite{laine2016temporal, sajjadi2016regularization, tarvainen2017mean} techniques mainly make the prediction on two views of one instance consistent. $\Pi$-Model~\cite{sajjadi2016regularization} focuses on reducing the distance of prediction between one instance and its stochastic perturbation. Unlike \textcolor{black}{the} $\Pi$-Model, temporal ensembling~\cite{laine2016temporal} adopts the ensemble of predictions as the target to achieve more stable performance, \textcolor{black}{while} Virtual Adversarial Training (VAT)~\cite{miyato2018virtual} explores adversarial disturbances of the unlabeled instances on the prediction of \textcolor{black}{the target} classifier. 
Pseudo-Labeled based approaches~\cite{berthelot2019remixmatch, berthelot2019mixmatch, lee2013pseudo, sohn2020fixmatch} annotate some unlabeled instances with pseudo labels to expand the labeled data. By leveraging the class probability of the unlabeled data, a pseudo-labeling method is proposed~\cite{lee2013pseudo}. Furthermore, FixMatch~\cite{sohn2020fixmatch} uses the weakly augmented unlabeled instances to create a pseudo label and enforce consistent prediction against its strong augmented version. 

\textcolor{black}{These traditional SSL approaches perform well when the class distribution is matched, but they suffer severe performance degradation under class distribution mismatch.}

\noindent {\bf Semi-Supervised Learning under Class Distribution Mismatch.} 
To tackle class distribution mismatch, several studies~\cite{28_SSL_USAD, 13_SSL_DS3L, ssl_huang2021trash, yang2022class} adopt the traditional SSL strategies with \textcolor{black}{the} assistance of soft or hard weights. UASD~\cite{28_SSL_USAD} leverages a threshold to the accumulated network's output to eliminate \textcolor{black}{the instances with} unknown categories, followed by pseudo-labeling highly confident \textcolor{black}{ones}. Similarly, T2T~\cite{ssl_huang2021trash} adopts a hard weight on the out-of-distribution score to conduct filtering and leverages consistency constraints to expand the labeled pool. 
\textcolor{black}{Furtherly, CCSSL~\cite{yang2022class} filters out unknown instances by taking both hard and soft weights into consideration.} 
These \textcolor{black}{approaches} with hard weights may eliminate too many instances from target categories.
\textcolor{black}{Instead of hard weights, }
 $\text{DS}^3\text{L}$~\cite{13_SSL_DS3L} assigns soft weights to unlabeled instances according to the consistent empirical risk loss.
However, SSL with pseudo labeling or consistency regularization heavily rely on the performance of the target classifier, 
\textcolor{black}{and thus they are susceptible to being invaded by instances with unknown categories.}

Additionally, a model-level \textcolor{black}{approach}~\cite{zhao2020robust} is proposed by modifying batch normalization to counter the unknown categories. Also, ORCA~\cite{ssl_cao2021open}, a novelty detection approach, leverages uncertainty-based adaptive margins to circumvent the bias caused by the mismatched distribution.

\noindent{\bf{Knowledge Distillation.}} 
Knowledge distillation aims to transfer knowledge from a big model  (teacher model) to a smaller one (student model)~\cite{wang2021knowledge}. It is widely applied to two distinct fields: model compression and knowledge transfer. 
Model compression is training a small student model to mimic the big teacher model or the ensemble of models. Buciluǎ et al.~\cite{bucilua2006model} compress the ensembles of the neural networks into a single one. While the approaches based on transfer knowledge concentrate more on effectively transferring and are mainly divided into logits-based and representation-based distillation~\cite{zhao2022decoupled}. The logits-based distillation approaches usually train the student model by leveraging the output of the teacher model as the soft label~\cite{wang2021knowledge}. Ba et al.~\cite{ba2014deep} propose to push the logits, i.e., the output before the softmax function, of the shallow neural network to mimic the ones from a deep neural network. Furtherly, Hinton et al.~\cite{hinton2015distilling} suggest training a student model to match the combination of the softmax distribution of the teacher model and ground truth. Representation-based approaches enable the student model to learn information from the intermediate layers~\cite{wang2021knowledge}. Kim et al.~\cite{kim2018attention} propose transferring the attention map from the teacher to the student. Park et al.~\cite{park2019relational}  introduce a novel approach that transfers the mutual relationship of the instances learned from the teacher to the student, similar to our intention.

However, these approaches mentioned above aim to transfer as much information as possible to the student model and ignore the unknown instances under class distribution mismatch,
\textcolor{black}{which} may \textcolor{black}{severely hurt} the training of the student model. 
Unlike the conventional approaches, WAD  is a weight-aware distillation framework that selectively transfers the knowledge to the student model, as shown in Figure~\ref{fig:x}, to fully use the beneficial knowledge and filter the \textcolor{black}{unknown ones} by weights. 
Specifically, WAD distills high-quality pseudo labels to the instances with target categories and filters the instances with unknown categories by assigning them tiny weights.

\section{Method}\label{sec:method}



In this section, we propose WAD, an SSL framework under class distribution mismatch. Concretely, Subsection~\ref{sbusec:problem} introduces the problem statement, followed by analyses of the SSL error in Subsection~\ref{sbusec:population}. Subsection~\ref{sbusec:framework} subsequently presents WAD. Finally, theoretical studies of WAD are conducted in Subsection~\ref{subsec:theoretical}.

\subsection{Problem Statement}\label{sbusec:problem}



In this study, we investigate the $K$ classification problem in an SSL manner \textcolor{black}{wherein limited labeled data $\mathcal{D}_l=\{(x_{i,l}, y_{i,l})\}_{i=1}^{m}$ and massive unlabeled instances $\mathcal{D}_u=\{x_{i,u}\}_{i=1}^{n}$ are accessible,  $x_{i,l} \in \mathcal{X}$, $y_{i,l} \in \mathcal{Y}$,  $\mathcal{Y}=\{1,..., K\}$ and $m \ll n$.} Under class distribution mismatch, the unlabeled instances are not guaranteed to belong to the $K$ target \textcolor{black}{categories} in $\mathcal{Y}$.


\subsection{Population Risk Analysis}\label{sbusec:population}

\begin{figure}[t]
	\centering
	\includegraphics[width=1\linewidth]{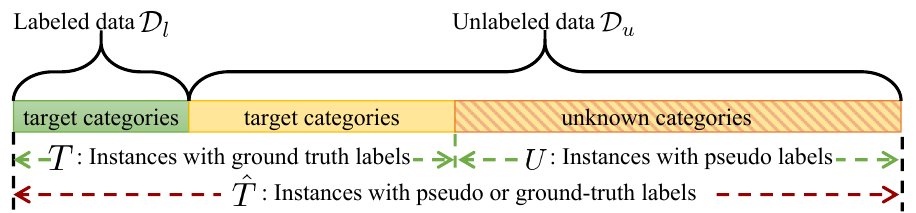}
	\caption{The  relations among data sets $\mathcal{D}_l$, $\mathcal{D}_u$, $T$, $U$, $\hat T$. \textcolor{black}{Note that $\hat T \neq T \cup U$ due to the instances with target categories in $\mathcal{D}_u$ are assigned with pseudo labels while not the ground truth ones in $T$. } }
	\label{fig:set}
\end{figure}

To make full use of the unlabeled data, we assign a pseudo label to each unlabeled instance, denoted as $\hat y$, and then build the target classifier, $h_{\hat T}:\mathcal{X}\rightarrow\mathcal{Y}$, to map the given instance to one of the known categories in $\mathcal{Y}$, where $\hat T = \{x_{i,l}, y_{i,l}\}_{i=1}^m \cup \{x_{i,u}, \hat y_{i,u}\}_{i=1}^n$, $\hat y_{i,u} \in \mathcal{Y}$. Here, $\hat T$ indicates the instances in hand, that is, labeled instances and unlabeled instances assigned with pseudo labels.
Then, the population risk~\cite{7diversity_coreset} of the target classifier learned from 
both labeled and unlabeled data with the pseudo label ($\hat T$) is controlled by the generalization gap, training error, and SSL error, as shown in~\myref{(1)}. The generalization gap is the gap between the population risk and the average 
 prediction loss across all instances with target categories ($T$).
\textcolor{black}{Note that $T$ contains all the accessible instances with target categories, including labeled and unlabeled. And every instance in $T$ is assumed with ground truth labels in ideal. }
The training error is the average empirical loss across $\hat T$. 
The SSL error is the gap between the average empirical loss across the instances \textcolor{black}{with target categories} ($T$) and the average empirical loss across \textcolor{black}{both} labeled data and unlabeled ones with pseudo labels ($\hat T$). We depict the relations among these sets in Figure~\ref{fig:set}. 


\begin{small}
\begin{equation}\label{(1)}
	\begin{aligned}
		&\mathbb{E}_{(\bm{x}, y) \sim D}[l(\bm{x}, y; h_{\hat T})]\\
		&\le \underbrace{\bigg| \mathbb{E}_{(\bm{x}, y) \sim\! D}[l(\bm{x}, y; h_{\hat T})] - \frac 1{|T|} \sum\limits_{(\bm{x}, y) \in T}l{(\bm{x}, y; h_{\hat T})} \bigg|}_{\mathbf{generalization\ gap}}\\
		&+  \underbrace{\bigg|\frac 1{|\hat T|} \sum\limits_{(\bm{x}, y) \in \hat T}l{(\bm{x}, y; h_{\hat T})}\bigg| }_{\mathbf{training\ error}}\\
		& +  \underbrace{\bigg|\frac 1{|T|} \sum\limits_{(\bm{x}, y) \in T}l{(\bm{x}, y; h_{\hat T})} -  \frac 1{|\hat T|} \sum\limits_{(\bm{x}, y) \in \hat T}l{(\bm{x}, y; h_{\hat T})}\bigg| }_\mathbf{SSL \ error},
	\end{aligned}
\end{equation}
\end{small}
where $\mathcal{D}$ is the data distribution of the instances that belong to target categories in the realistic world, i.e., $\mathcal{D}=\mathcal{X} \times \mathcal{Y}$. $l(\cdot, \cdot; h_{\hat T}): \mathcal{X} \times \mathcal{Y} \rightarrow \mathcal{R} $ denotes the loss function of the classifier $h_{\hat T}$ learned from $\hat T$. 


Theoretical analyses~\cite{41_theoretical} have confirmed that the generalization gap of DNNs can be bounded, and \textcolor{black}{empirical evidence suggests that the training error of DNNs can be reduced almost to zero~\cite{7diversity_coreset}.}
Thus, the essential component concerning population risk is the SSL error. 
Under class distribution mismatch, in addition to the wrongly annotated instances with target categories, the ones with unknown categories also contribute to the SSL error as they invade the training of the target classifier as outliers. 
Accordingly, we decouple the SSL error into pseudo-labeling and invasion error, as shown in~\myref{(2)}. For a detailed derivation process, please \textcolor{black}{refer to Appendix 5.2.}
	\begin{align}\label{(2)}
		& \bigg|\frac 1{|T|} \sum\limits_{(\bm{x}, y) \in T}l{(\bm{x}, y; h_{\hat T})} -  \frac 1{|\hat T|} \sum\limits_{(\bm{x}, y) \in \hat T}l{(\bm{x}, y; h_{\hat T})}\bigg| \nonumber \\
		& \le \underbrace{ \bigg|\frac 1{|T|} \sum\limits_{(\bm{x}, y) \in T}l{(\bm{x}, y; h_{\hat T})}  - \frac 1{|\hat T|} \sum\limits_{(\bm{x}, y) \in \hat T \backslash U} l{(\bm{x}, y; h_{\hat T})}\bigg|}_{\mathbf{Pseudo-labeling \ error}} \nonumber \\
		&+  \underbrace{ \bigg|\frac 1{|\hat T|} \sum\limits_{(\bm{x}, y) \in U} l{(\bm{x}, y; h_{\hat T})}\bigg|}_{\mathbf{Invasion\ error}}
	\end{align}
where $U$ indicates the unlabeled instances with unknown categories and $\hat T = \hat T \backslash U \cup U$.


In ~\myref{(2)}, the pseudo-labeling error is contributed by the wrongly annotated instances with target categories, as it is the gap of the average empirical loss caused by the inconsistency of the ground truth and pseudo labels. Thus, the quality of pseudo-labels assigned to unlabeled instances within the target distribution determines this error, and accurate pseudo-labeling may alleviate it. By contrast, the invasion error is the average empirical loss across the instances \textcolor{black}{with} unknown categories that is caused by the negative effect of those untargeted instances. By ~\myref{(1)}\&~\myref{(2)}, 
	\textcolor{black}{we find} that the population risk of the target model is jointly controlled by pseudo-labeling error and invasion error. Accordingly, to mitigate the SSL error, we need to filter those instances with unknown categories and accurately annotate the unlabeled instances with target categories as well.


\subsection{\textcolor{black}{Weight-aware Distillation Framework}}\label{sbusec:framework}

\begin{figure*}[t]
	\centering
	\includegraphics[width=1\linewidth, height=6.5cm]{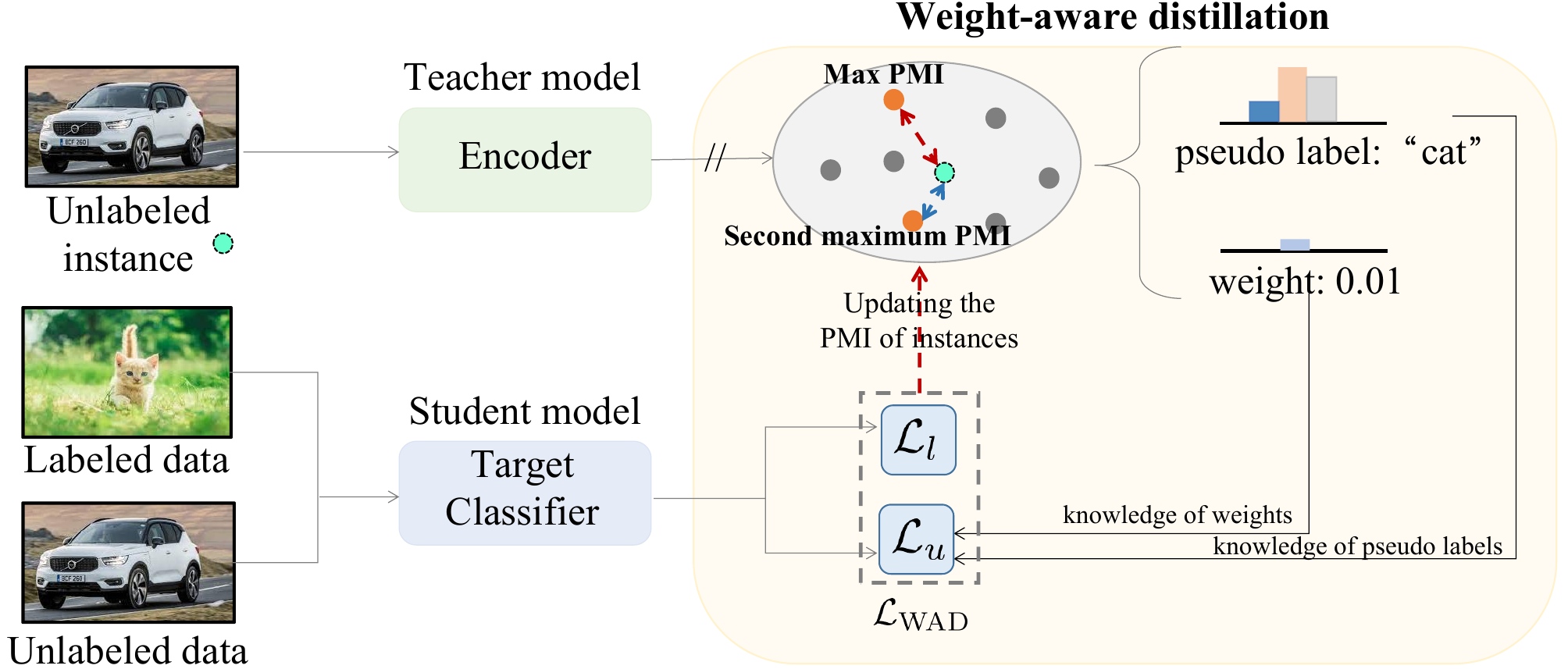}
	\caption{Illustration of WAD. The pseudo labels and weights determined by point mutual information (PMI) in robust representations of the teacher model are used to participate in training the target classifier. Then, some reliable instances selected according to  $\mathcal{L}_{\text{WAD}}$ are regarded as labeled ones to update the knowledge gradually.  ``\textcolor{orange}{$\bullet$}'' and ``\textcolor{gray}{$\bullet$}'' indicates the labeled and unlabeled instance respectively. ``$//$'' means stop gradient.}
	\label{fig:framework}
\end{figure*}


With the aim of mitigating the pseudo-labeling and invasion errors, we design \textcolor{black}{an  SSL framework named WAD, which} delivers the knowledge of pseudo labels and weights from robust representations to the target classifier. 

\subsubsection{Pseudo Label Learning}\label{subsec:pse}

Most existing SSL approaches produce pseudo labels by leveraging an immature target classifier, which cause catastrophic error once invaded by some instances with unknown categories, just as discussed in Section~\ref{sec:intro}. 
To solve this problem, we distill the pseudo labels from a representation space (Teacher model) which is learned from all labeled and unlabeled instances by contrastive learning in an unsupervised manner and then transfer it to the target classifier (Student model). The teacher model could produce closely aligned representations for instances from the same categories and maximize the mutual information among them~\cite{NCE_ref2, infoNCE, 29_contrastive_learning_survey,57_contrastive_review}. 

Denoted the labeled and unlabeled representations learned by the teacher model, $\phi$, as $\mathcal{Z}_{l}=\{z_{j,l,k}\}_{j=1}^{m}$  and $\mathcal{Z}_u=\{z_{i,u}\}_{i=1}^{n}$, respectively, where $k \in \mathcal{Y}$.
Inspired by the characteristic of contrastive learning, one effective approach for building the pseudo-label of an unlabeled instance is to identify the labeled instance with the highest PMI and then assign the label of it to unlabeled ones. The PMI between the unlabeled and labeled representation is formulated as~\myref{pmi}.
\begin{equation}\label{pmi}
	\begin{aligned}
		\text{PMI}(z_{i,u}, z_{j,l,k}) = log\left[\frac{p(z_{j,l,k}|z_{i,u})}{p(z_{j,l,k})}\right]
	\end{aligned}
\end{equation}
Although the conditional and marginals distributions, i.e., $p(z_{j,l,k}|z_{i,u})$ and $p(z_{j,l,k})$, cannot be directly evaluated, we prove that PMI is proportional to the inner product \textcolor{black}{in Appendix 2,} as described in~\myref{(P)}.
\begin{equation}\label{(P)}
	\begin{aligned}
		f(z_{i,u}, z_{j,l,k}) \propto \text{PMI}(z_{i,u}, z_{j,l,k})
	\end{aligned}.
\end{equation}
where $f=cos(\cdot,\cdot)$, $\Vert z_{i} \Vert=1$, and $\propto$ stands for ``proportional to''.

Therefore, the pseudo label is formulated as~\myref{pesudo}.
\begin{equation}\label{pesudo}
	\begin{aligned}
		\hat y_{i,u} = \mathop{\arg\max}_{k} f(z_{i,u}, z_{j,l,k})
	\end{aligned}
\end{equation}
Consequently, the class label of the labeled instance with the maximum PMI is assigned to the unlabeled one.
The~\myref{pesudo}  can precisely capture the PMI from the representation space to produce high-quality pseudo labels and 
then mitigate pseudo-labeling error.

\subsubsection{Unknown Categories Filtering}



To mitigate the invasion error, the instances with unknown categories should be filtered out. Following the Subsubsection~\ref{subsec:pse}, a higher PMI between the labeled and unlabeled instance suggests a stronger association or similarity between the two instances, further indicating a higher likelihood of the unlabeled instance belonging to the same class distribution as the labeled one. However, some hard instances that have similar PMI between two target categories, i.e., laid on the decision boundary of two target categories, may introduce incorrect pseudo labels and hurt the performance of the target classifier. Hence, we also propose a ratio among the first and second maximum PMI to evaluate the confidence of the pseudo labels. Then, the weight is defined as~\myref{weight} to avoid the negative effect caused by the wrong labels and unknown categories. 
\begin{equation}\label{weight}
\begin{aligned}
	\bm{w}_{i,u} = g_1\left(\widetilde{p}_{i,u}\right)\times g_2\left(1-\frac{ \widetilde{q}_{i,u}}{\widetilde{p}_{i,u}}\right), \\
\end{aligned}
\end{equation}
\textcolor{black}{wherein, } 
\begin{equation*}
\begin{aligned}
	&\widetilde p_{i,u} = \mathop{\max}_{j} f(z_{i,u}, z_{j,l,k}) \\
	&\widetilde q_{i,u} =  \mathop{\max}_{v, k\neq \hat y_{i,u}} f(z_{i,u}, z_{v,l,k}) 
\end{aligned}
\end{equation*}
In~\myref{weight}, $g_1(\cdot)$ and $g_2(\cdot)$ can be interpreted as any monotonically increasing functions.  
The former in~\myref{weight} aims to estimate the likelihood of the unlabeled instance belonging to target categories. The higher this item, the more chances of the instance in the target class distribution are. The latter, $g_2(\cdot)$, penalizes instances whose labels are ambiguous between the nearest and second-nearest target categories. The lower this item, the larger probability of incorrect pseudo labels is. As shown in Figure~\ref{fig:information-updated},  the weight could filter instances with unknown categories and those incorrectly annotated ones with target categories, while the ones from target categories with high-quality pseudo labels are encouraged. \textcolor{black}{Thus, by weight, WAD selectively distills the knowledge beneficial to the target classifier from the teacher model, and the invasion error is then mitigated.}

\subsubsection{Weight-aware Knowledge Distillation}
{\bf{Weight-aware knowledge distillation loss.}} 
The knowledge of pseudo labels and weights captured from robust representations is applied in the distillation process. In each feed-forward process, pseudo labels and weights are aggregated to the target classifier, as shown in Figure~\ref{fig:framework}. Then, we propose the weight-aware knowledge distillation loss, including the traditional supervised loss $\mathcal{L}_l$ in labeled data and weight-aware supervised loss $\mathcal{L}_u$ in unlabeled data as ~\myref{loss}.
\begin{equation}\label{loss}
	\begin{aligned}
		\mathcal{L}_{\text{WAD}} &= \mathcal{L}_l + \mathcal{L}_u,
	\end{aligned}
\end{equation}
wherein,

\begin{equation*}
	\begin{aligned}
		&\mathcal{L}_l = \frac{1}{|\mathcal{D}_{l}|} 	
		\mathop{\sum}\limits_{(x_{i,l},  y_{i,l}) \in \mathcal{D}_{l}} \ell(h(x_{i,l};\theta),  y_{i,l})\\
		&\mathcal{L}_u = \frac{1}{|\mathcal{D}_u|} \mathop{\sum}\limits_{x_{i,u}\in\mathcal{D}_u} \bm{w}_{i,u} \ell(h(x_{i,u};\theta), \hat y_{i,u})
	\end{aligned}
\end{equation*}

The traditional supervised loss $\mathcal{L}_l$ aims to minimize the distance between the predicted probability and the ground truth label. While the weight-aware supervised loss $\mathcal{L}_u$ mainly focuses on \textcolor{black}{selectively} transferring the beneficial knowledge from the teacher model to the student model \textcolor{black}{by weights} to mitigate the negative effect from unknown categories and improve the target classifier as well. 
Moreover, $\mathcal{L}_{\text{WAD}}$ is the loss function that is adopted to train the target classifier $h_{\hat T}$ mentioned in~\myref{(1)}. Consequently, WAD leverages the pseudo labels and weights to mitigate the pseudo-labeling and invasion errors, following alleviating the SSL error, which has been proved in Subsection~\ref{subsec:theoretical}.

\noindent{\bf{Knowledge-update in Training.}} 
The knowledge of pseudo labels and weights may be biased as the labeled data is limited. Accordingly, after several forward iterations, we progressively add some reliable instances to labeled data. Because the feedback from the target classifier, i.e., loss, is highly related to the weights and reflects the training error, we consider the reliability according to it. Then,  the criterion for updating is formulated as~\myref{pse_true}. 
 \begin{equation}\label{pse_true}
 	\begin{aligned}
 		c_{i,u} = \ell(h(x_{i,u};{\theta}_t), \hat y_{i,u})
 	\end{aligned}
 \end{equation}
 where $\ell(\cdot, \cdot)$ is the cross-entropy function, and ${\theta}_t$ is the parameters of the target classifier in the current iteration.
 
\begin{figure*}[t]
	\centering
	\includegraphics[width=1\linewidth]{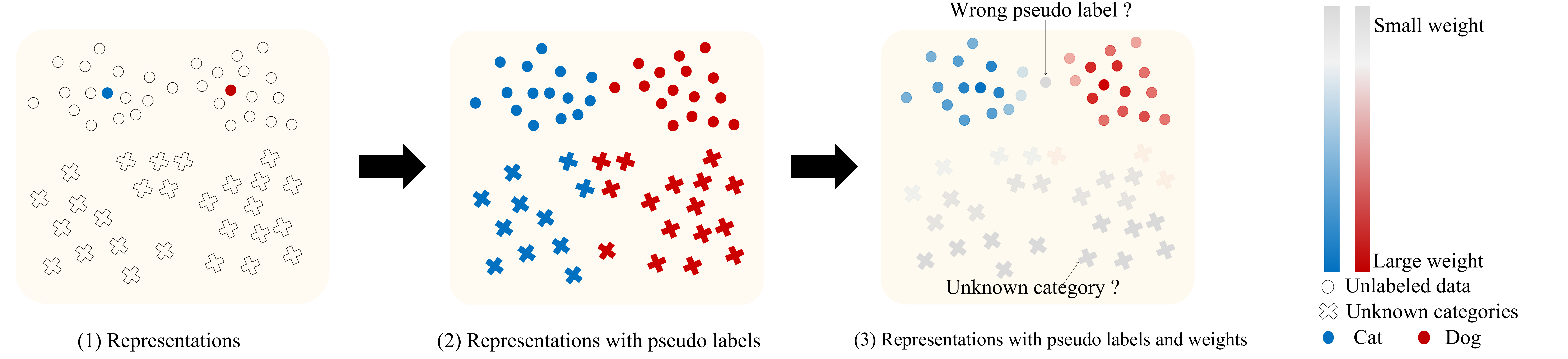}
	\caption{\textcolor{black}{The schematic diagram of how WAD works. First, each unlabeled instance is assigned a pseudo label according to the max PMI with labeled data in (1), as shown in (2). Then, the instances with wrong labels and unknown categories are assigned a tiny weight to avoid the negative effect, as shown in (3). }}
	\label{fig:information-updated}
\end{figure*}

The reliability of $x_{i,u}$ is enhanced when $c_{i,u}$ takes a lower value. Then, WAD leverages~\myref{pse_true} to identify the top $\alpha\%$ reliable instances from the unlabeled data and puts them in the labeled data while removing them from the unlabeled data. Moreover, we adopt the polynomial decay~\cite{borichev2010optimal} to dynamically adjust $\alpha$ to prevent the gradually increased negative effect from unknown categories with the iteration. The details are shown \textcolor{black}{in Appendix 3.} A visualization of the number of selected reliable instances with target categories is also provided \textcolor{black}{in Appendix 4.4.} \textcolor{black}{Consequently,} the pseudo labels and weights are updated in the \textcolor{black}{subsequent} distillation steps, as shown in Figure~\ref{fig:framework}, with the aim of optimizing the target classifier. Finally, the schematic diagram and algorithm process is presented in Figure~\ref{fig:information-updated} and Algorithm~\ref{algorithm}, respectively.

\subsection{Theoretical Studies}\label{subsec:theoretical}

This subsection provides the theoretical studies about the WAD's SSL error, as shown in Theorem \ref{theorem}. 
Detailed proof of Theorem \ref{theorem} is given in \textcolor{black}{Appendix 5.}
\begin{theorem}\label{theorem}
Given $|T|$ instances that i.i.d. sampled from  $\mathcal{D}$ as $\{ (\bm{x}_i, y_i) \}_{i=1}^{|T|}$, $|U|$ instances that is not i.i.d with  $\mathcal{D}$, and $\hat T = \{(x,\hat y)| (x,y)\in T \cup U, \hat y \in \mathcal{Y}\}$ where $\mathcal{D}=\mathcal{X}\times\mathcal{Y}$. 
 Assume the loss function $l(\cdot, y; {h_{\hat T}})$ is $\lambda^{l}$-Lipschitz continuous for all $y, h_{\hat T}$ and bounded by $H$, the regression function is $\lambda^{\mu}$-Lipschitz continuous, training error $l(\bm{x}, y; {h_{\hat T}})=0$ , $\forall (\bm{x}, y) \in \hat T$. $\overline{\bm{w}}$ indicates the average of weights, and 
 $\xi$ is the maximum PMI which \textcolor{black}{determines} the pseudo label, with the probability of at least $1-\gamma$,
 	\small
 	\begin{align}
 		&\bigg|\frac 1{|T|} \sum\limits_{(\bm{x}, y) \in T}l{(\bm{x}, y; h_{\hat T})}  - \frac 1{|\hat T|} \sum\limits_{(\bm{x}, y) \in \hat T \backslash U} l{(\bm{x}, y; h_{\hat T})}\bigg| \nonumber\\
 		&+ \bigg|\frac 1{|\hat T|} \sum\limits_{(\bm{x}, y) \in U}l{(\bm{x}, y; h_{\hat T})} \bigg| \\
 		&\le \sqrt{4-4\xi}(\lambda^l + \lambda^{\mu}HK) + \frac{\overline{\bm{w}}|U|H}{|\hat T|} +  \sqrt{\frac{2H^2log(1/\gamma)}{|T|}}.\nonumber
 	\end{align}
\end{theorem}

From Theorem \ref{theorem}, we find that the smaller $\overline{\bm{w}}$ and the larger  $\xi$, the tighter the bound in the SSL error is. Specifically, just as verified in Appendix 5, the pseudo-labeling 
error bounded by 
$\sqrt{4-4\xi}(\lambda^l + \lambda^{\mu}HK) + \sqrt{\frac{2H^2log(1/\gamma)}{|T|}}$ and the invasion error bounded by $\frac{\overline{\bm{w}}|U|H}{|\hat T|} $ can be reduced by minimizing the weights $w$ of unlabeled instances with unknown categories and  maximizing the confidence of pseudo labels, just as WAD does. Thus, WAD's SSL error has a tight upper bound.

 \begin{algorithm}[h]
	\label{algorithm}	
	\caption{\textcolor{black}{Weight-aware distillation framework}}
	\KwIn{
		Labeled data: $\mathcal{D}_l$, Unlabeled data: $\mathcal{D}_u$, 
		Max iterations: $N$, Initial value of $\alpha$: ${\alpha}_{0}$, Set of update steps: $G$.}
	
	\KwOut{target classifier $h_{\hat T}$.}
	
	Embedding:  $\mathcal{Z}_{l} \leftarrow \phi{(\mathcal{D}_l)}$, $\mathcal{Z}_u \leftarrow \phi{(\mathcal{D}_u)}$. 
	
	Initialize model parameter;\
	
	\For{$t=0$ to $N-1$ }{
		
		$\forall \bm{x}_{i,u} \in \mathcal{D}_u$, distill the knowledge about pseudo label, $\hat y_{i,u}$,  using~\myref{pesudo};\\
		$\forall \bm{x}_{i,u} \in \mathcal{D}_u$, distill the knowledge about weight, $\bm{w}_{i,u}$,  using~\myref{weight};\\
		$\alpha$ = polynomial\_decay(${\alpha}_0$);\\	
		Using~\myref{loss} training the target classifier;\\	
		\uIf{$t \in G$}{
			$C=\emptyset$;\\
			Calculate the reliability of each unlabeled instance using~\myref{pse_true}.\\
			Ascending the instances according to reliability and add the top $\alpha\%$ instances with pseudo labels to $C$.\\
			$\mathcal{D}_l = \mathcal{D}_l \cup C$,
			$\mathcal{D}_u = \mathcal{D}_u \backslash \{x|(x,y) \in C\}$.}
	}
	{\bf{Return}} target classifier $h_{\hat T}$.\
\end{algorithm}

\section{Experiments}\label{sec:exper}

\begin{table*}[ht]
	\small
	\centering
	{\begin{minipage}{1\linewidth}
			\centering
			\resizebox{1\textwidth}{!}{
				\setlength{\tabcolsep}{1.3mm}{
					\begin{tabular}{lcccccccccc}
						\toprule
						
						& & \multicolumn{4}{c}{CIFAR10} & & \multicolumn{4}{c}{CIFAR100} \\
						\cline{1-1} \cline{3-6}\cline{8-11}
						Method & & 20\% & 40\% & 60\% & 80\% & & 20\% & 40\% & 60\% & 80\% \\
						\cline{1-1} \cline{3-6}\cline{8-11}
						Baseline & & 94.33$\pm$0.45 & 94.33$\pm$0.45 & 94.33$\pm$0.45 & 94.33$\pm$0.45 & & 36.98$\pm$1.79 & 36.98$\pm$1.79 & 36.98$\pm$1.79 & 36.98$\pm$1.79\\
						$\text{DS}^3\text{L}$ & & 91.82$\pm$1.89 & 91.38$\pm$1.73 & 92.47$\pm$0.25 & 90.82$\pm$1.50 & & 23.92$\pm$2.78 & 24.92$\pm$4.41 & 26.20$\pm$4.29 & 24.55$\pm$3.67\\
						UASD & &  95.02$\pm$0.77 & 95.03$\pm$0.77 & 93.87$\pm$0.13 & 93.37$\pm$0.35 &  &  39.85$\pm$0.35 & 37.55$\pm$2.24 & 36.03$\pm$0.73 & 29.87$\pm$2.07  \\
						CCSSL & & 86.08$\pm$0.12 & 84.00$\pm$0.17 & 83.13$\pm$0.19 & 81.15$\pm$0.25 & & 41.72 $\pm$0.85 & 41.20$\pm$0.58 & 40.60 $\pm$0.22 & 39.67$\pm$0.31\\
						T2T & & - & - & - & - & & 43.70$\pm$0.50 & 42.82$\pm$0.45 & 40.12$\pm$0.71 & 37.35$\pm$1.10\\
						T2T$\bf{w}\backslash o$ pre. & & - & - & - & - & & 39.40$\pm$0.36 & 36.78$\pm$0.16 & 36.65 $\pm$1.09 & 34.62$\pm$1.68\\
						ORCA & & 95.40$\pm$0.74 & 94.13$\pm$1.16 & 94.35$\pm$0.67 & 93.82$\pm$0.93 & & 29.50$\pm$0.25 & 31.12$\pm$0.71 & 31.18$\pm$0.40 & 31.65$\pm$1.86\\
						ORCA $\bf{w}\backslash o$ pre. & & 93.32$\pm$0.99 & 92.55$\pm$2.02 & 92.37$\pm$0.90 & 89.65$\pm$6.95 & & 22.13$\pm$1.33 & 23.98$\pm$0.79 & 23.37$\pm$1.14 & 22.98$\pm$0.53\\
						WAD & & \bf{98.43$\pm$0.14} & \bf{97.88$\pm$0.33} & \bf{97.90$\pm$0.20}  & \bf{97.77$\pm$0.33} & & \bf{51.65$\pm$2.86} & \bf{50.00$\pm$1.43} & \bf{46.88$\pm$0.20} & \bf{45.45$\pm$1.73}\\
						\bottomrule
				\end{tabular}}
			}
	\end{minipage}}
	\caption{Experimental results on CIFAR10 and CIFAR100 under different mismatch proportions.}\label{results_10_100}
\end{table*}


Subsection~\ref{sec:results} presents the comparison results between WAD and five state-of-the-art SSL \textcolor{black}{approaches}, as well as one standard baseline. Furthermore, an ablation experiment is conducted in Subsection~\ref{sec:ablation}, while sensitivity analyses and visualization are carried out in Subsection~\ref{sec:sensitivity} and Subsection~\ref{sec:visualization}, respectively. For more experiments, please refer to Appendix 4.2 \& 4.5.

\subsection{Experimental Setups}

{\bf{Datasets.}} Our experiments are conducted on two benchmark datasets, CIFAR10~\cite{cifar10-100} and CIFAR100~\cite{cifar10-100}, as well as an artificial cross-dataset that comprises subsamples from CIFAR10, CIFAR100, Flowers~\cite{flowers}, Food-101~\cite{food-101}, and Places-365~\cite{places-365}. The CIFAR10 and CIFAR100 datasets 
consist of 50,000 training and 10,000 testing images of 10 and 100 categories, respectively. The cross-dataset contains 138,000 unlabeled instances from 674 categories. 
All images from the datasets are resized to 32$\times$32. For further details, \textcolor{black}{please refer to Appendix 4.1}.

{\bf{Settings.}} \romannumeral1 ) The proportion of the instances with unknown categories in unlabeled data, named as mismatch proportion, are set as 20\%, 40\%, 60\%, and 80\% in this work. For instance, the unlabeled data has a 60\% mismatch proportion with 4,000 instances with target categories and 6,000 instances with unknown categories. 
 \romannumeral2 ) Randomly sampled 8\% instances from the training dataset that belong to target categories are regarded as labeled data. The remaining 92\% of instances with target categories and some instances with unknown categories are composed of unlabeled data according to the mismatch proportion.
 


{\bf{Details.}} The teacher model is with a Resnet-18~\cite{resnet} backbone and is trained using SimCLR~\cite{30_contrastive_learning_simclr}. And we maintain consistency with SimCLR in all implementation details. The target  classifier is a WideResnet-28-2 network~\cite{wide_resnet} with input size $32 \times 32$, following Huang et al.~\cite{ssl_huang2021trash}. \textcolor{black}{Both the encoder and target classifier are trained from scratch.} We train the target classifier using the Adam optimizer~\cite{45_adam} with a learning rate of $5\times 10^{-4}$. Furthermore, the epochs and batch size are set as 100 and 32, respectively. The augmentations include random horizontal flipping, random translation by up to 2 pixels, and Gaussian input noise with a standard deviation of 0.15 is used in the training of the target classifier as same as Guo et al.~\cite{13_SSL_DS3L}. Moreover, we apply global contrast normalization and ZCA normalized, which is widely used in the pretreatment~\cite{13_SSL_DS3L, 28_SSL_USAD}, on CIFAR10. For simplicity, the functions $g_1(\cdot)$ and $g_2(\cdot)$ act as identical mapping with no additional constraints. The initial value of $\alpha$  is set as 0.1 and decayed five times until it reached 0. It remains the same across all experiments unless otherwise specified. Finally, the approaches on each dataset run three times, and the mean accuracy and standard deviation are reported; the best one is highlighted in bold.



{\bf{Baselines.}} WAD is compared to five state-of-the-art approaches, including $\text{DS}^3\text{L}$~\cite{13_SSL_DS3L}, T2T~\cite{ssl_huang2021trash}, CCSSL~\cite{yang2022class}, UASD~\cite{28_SSL_USAD} and ORCA~\cite{ssl_cao2021open}, as well as one baseline model that only trains labeled data. Moreover, T2T and ORCA are performed without pretraining tasks for fairness, indicated by ``T2T {\bf{w$\backslash$o}} pre.'' and ``ORCA {\bf{w$\backslash$o}} pre.'' .
\subsection{Experimental Results}\label{sec:results}


This subsection presents the experimental results of the classification tasks performed on CIFAR10, CIFAR100, and a cross-dataset. For CIFAR10, we designated two categories as the target and eight as unknown, while twenty classes are considered as target categories and eighty categories as unknown in CIFAR100. Moreover, we constructed a cross-dataset integrated with five datasets to evaluate WAD in the case that the unlabeled data contains massive unknown categories. Specifically, six classes from CIFAR10 were assigned as target categories, and 668 categories from four external datasets are unknown. The experimental results conducted on CIFAR10, CIFAR100, and the cross-dataset are presented in Table~\ref{results_10_100} and Table~\ref{results_cross}. 

\noindent{\bf{Results on CIFAR10 and CIFAR100.}} From Table~\ref{results_10_100}, we have four findings as follows. 
\romannumeral1 ) WAD outperforms all compared methods on CIFAR10 and CIFAR100 with different mismatch proportions, demonstrating its remarkable performance.
\romannumeral2 ) WAD retains stable performance improvement under different mismatch proportions, exhibiting further improvement of 4.1\%, 3.55\%, 2.91\%, and 3.44\% for mismatch proportions of 20\%, 40\%, 60\%, and 80\% on CIFAR10. This highlights that WAD can achieve 
robust performance even under a high mismatch proportion. 
\romannumeral3 ) The accuracies of $\text{DS}^3\text{L}$ on CIFAR10 and CIFAR100 are lower than baseline, as ORCA does. This is because it weights the instances according to consistent empirical risk loss, resulting in the invasion of many unknown categories in training, \textcolor{black}{as shown in Appendix 4.3.} 
\romannumeral4 ) In CIFAR100, WAD surpasses baseline  8.47\% for 80\% mismatch proportion. This demonstrates that WAD is still effective when the unlabeled data contains large unknown categories. 
Therefore, WAD achieves outstanding performance on datasets with different mismatch proportions and exhibits excellent robustness to the scale of unknown categories. 
Notably, T2T can not apply to the binary classification task, and the accuracy is not reported here.

\begin{table}[t]
	\small
	\centering
	{\begin{minipage}{1\linewidth}
			\centering
			\resizebox{1\textwidth}{!}{
				\setlength{\tabcolsep}{0.4mm}{
					\begin{tabular}{lcccccccccc}
						\toprule
						
						& & \multicolumn{4}{c}{Cross-dataset}  \\
						\cline{1-1} \cline{3-6}
						Method & & 20\% & 40\% & 60\% & 80\% \\
						\cline{1-1} \cline{3-6}
						Baseline & & 66.83$\pm$1.37 & 66.83$\pm$1.37 & 66.83$\pm$1.37 & 66.83$\pm$1.37 \\
						$\text{DS}^3\text{L}$ & & 50.02$\pm$6.69 & 50.69$\pm$5.26 & 49.03$\pm$5.93 & 51.46$\pm$6.99 \\
						UASD & & 61.18$\pm$0.29 & 57.02$\pm$0.58 & 54.70$\pm$2.25 & 45.67$\pm$1.72  \\
						CCSSL & & 64.83$\pm$0.27 & 65.15$\pm$0.56 & 64.16$\pm$0.58 & 64.16$\pm$0.45 \\
						T2T & & 66.56$\pm$2.80 & 65.08$\pm$0.76 & 63.76$\pm$0.53 & 62.83$\pm$0.77 \\
						T2T $\bf{w}\backslash o$ pre. & & 64.44$\pm$0.15 & 62.47$\pm$0.79 & 62.23$\pm$0.75 & 61.42 $\pm$0.65 \\
						ORCA & & 65.53$\pm$0.85 & 65.51$\pm$1.25 & 66.44$\pm$0.80 & 66.46$\pm$1.28 \\
						ORCA $\bf{w}\backslash o$ pre. & & 65.37$\pm$0.78 & 63.63$\pm$0.64 & 64.42$\pm$0.53 & 66.34$\pm$1.05 \\
						WAD & & \bf{67.13$\pm$0.59} & \bf{67.20}$\pm$1.65 & \bf{67.80$\pm$0.07} & {\bf{67.88$\pm$0.37}} \\
						\bottomrule
					\end{tabular}
			}}
	\end{minipage}}
	\caption{Experimental results on cross-dataset under different mismatch proportions.}\label{results_cross}
\end{table}

\noindent{\bf{Results on cross-dataset.}} Further, we investigate the limits of WAD's tolerance for unknown categories and then perform the experiments on an artificial cross-dataset containing 668 unknown categories from four datasets. From Table~\ref{results_cross},
we observe that WAD still maintains an improvement compared to the baseline. Obviously, the other compared methods were lower than the baseline. This indicates that WAD could boost the
performance even on a dataset that contains 
massive instances with unknown categories.

\subsection{Ablation Studies}\label{sec:ablation}

We conducted ablation studies on the CIFAR10 dataset using different models: "+Pse." (trained with labeled data and unlabeled instances with pseudo labels), "+Pse.\&W." (trained with pseudo labels and fix weights), and the WAD model (trained with all components). We also examined the impact of the weight function and explored alternative choices for $g(\cdot)$ through identical mappings, $g_i(\cdot)$, and the transformation $\tilde{g_i}(\cdot)=exp(\cdot)$. Results are presented in Table~\ref{ablation}, and w$\backslash$o $g_i(\cdot)$ means removing $g_i(\cdot)$ from~\myref{weight}.


\noindent{\bf{Effects of pseudo labels.}} From Table~\ref{ablation}, we observe that compared with the baseline, 2.72\% and 1.52\%  accuracy improvement can be obtained by leveraging the unlabeled instances with pseudo labels, under 20\% and 80\% mismatch proportion, respectively. This indicates that the pseudo labels are beneficial to improving performance.

\noindent{\bf{Effects of weights. }} According to Table~\ref{ablation}, we observe two findings about weights. \romannumeral1 ) Training by leveraging both pseudo labels and weights exhibits the comparable performance to that without weights under 20\% mismatch proportion. This is because there are fewer instances with unknown categories under 20\%. Then, the model training with fixed weights will result in a sub-optimal model compared to explicit labels. \romannumeral2 ) The weights improve the accuracy by 0.99\% over without it, under 40\% mismatch proportion, while the gap decreases with the mismatch proportion increasing. 
This demonstrates that the weights are effective in filtering the instances with unknown categories. And the performance degradation is because the absence of the knowledge-update makes the algorithm fail to prevent the increased negative effect from unknown categories.


\noindent{\bf{Effects of knowledge-update.}} We have two findings according to Table~\ref{ablation}. \romannumeral1 ) The accuracy with knowledge-update surpasses the one only leveraging pseudo labels and weights, and the gap between them reaches 1.77\% under 80\% mismatch proportion. This indicates that 
knowledge-update
 plays important roles in WAD. \romannumeral2 ) WAD, training with all components, shows its outstanding performance compared to the ones removing other parts. This demonstrates that the aggregation of all the proposed parts could achieve significant improvement.
 
 \noindent{\bf{Effects of each part of~\myref{weight}.}}
 From the Table~\ref{ablation}, we have the following two findings.
\romannumeral1 ) both ``w$\backslash$o  $g_1(\cdot)$'' and ``w$\backslash$o $g_2(\cdot)$'' are worse than WAD, illustrating their equal importance for WAD. 
\romannumeral2 ) Assigning the same mappings to $g_1(\cdot)$ and $g_2(\cdot)$ yields better performance, as the same mappings share the same scales.
\begin{table}[h]
	\small
	\centering
	{\begin{minipage}{1\linewidth}
			\centering
			\resizebox{1\textwidth}{!}{
			\setlength{\tabcolsep}{0.8mm}{
				\begin{tabular}{lcccccccccc}
					\toprule
					Setting & 20\% & 40\% & 60\% & 80\%\\
					\midrule
					Baseline &94.33$\pm$0.45 &94.33$\pm$0.40 &94.33$\pm$0.4 & 94.33$\pm$0.45 \\
					+Pse. & 97.05$\pm$0.48& 95.98$\pm$0.75& 96.65$\pm$0.35& 95.85$\pm$0.88 \\
					+Pse.$\&$W & 96.62$\pm$0.47& 96.97$\pm$0.78& 97.22$\pm$0.38& 96.00$\pm$0.48 \\
					w$\backslash$o $g_1(\cdot)$ & 97.85$\pm$0.57 & 96.98$\pm$0.11 & 94.38$\pm$1.52 & 94.38$\pm$0.74 \\
					w$\backslash$o $g_2(\cdot)$ & 97.98$\pm$0.53 & 96.85$\pm$0.14 & 94.58$\pm$0.46 & 95.85$\pm$0.35 \\
					$\tilde{g_1}(\cdot) \times g_2(\cdot)$& 97.50$\pm$0.07 & 97.15$\pm$0.71 & 94.95$\pm$0.14  &  94.48$\pm$0.46 \\
					${g_1}(\cdot) \times \tilde{g_2}(\cdot)$& 96.60$\pm$0.57 & 96.93$\pm$0.04 & 95.65$\pm$0.21  &  95.65$\pm$0.07 \\
					WAD & \bf{98.43$\pm$0.14} & \bf{97.88$\pm$0.33} & \bf{97.90$\pm$0.20} & \bf{97.77$\pm$0.33}\\
					\bottomrule
				\end{tabular}
			}}
	\end{minipage}}
\caption{Ablation Studies under different mismatch proportions.}\label{ablation}
\end{table}

\subsection{Sensitivity Analysis}\label{sec:sensitivity}

This subsection investigates the influence of parameter $\alpha$, which controls how many instances with high reliability will be added to labeled data. Hence, we vary the initial value of $\alpha$ and evaluate WAD's performance on CIFAR10. The results are reported in Table~\ref{sensitive}. We find that WAD depicts the comparable performance with different values of $\alpha$, although lower values of $\alpha$ achieve slightly better performance with 20\% and 40\% mismatch proportions. This indicates that WAD is not sensitive to $\alpha$ because \textcolor{black}{the selected} instances may have higher similarities. Thus, WAD can achieve a robust performance for a wide range of $\alpha$ but not too large, preventing the invasion of unknown categories.


%
%

\begin{table}[h]
	\small
	\centering
	{\begin{minipage}{1\linewidth}
			\centering
			\resizebox{1\textwidth}{!}{
			\setlength{\tabcolsep}{0.6mm}{
			\begin{tabular}{l cccc}
				\toprule
				Setting & 20\% & 40\% & 60\% & 80\%\\
				\midrule
				$\alpha = 0.05$& \bf{98.62$\pm$0.28}& \bf{98.12$\pm$0.39}& 97.63$\pm$0.03& 97.63$\pm$0.19 \\
				$\alpha = 0.10$&98.43$\pm$0.14 &  97.88$\pm$0.33 & 97.90$\pm$0.20  & \bf{97.77$\pm$0.33}\\
				$\alpha = 0.15$& 98.35$\pm$0.25 & 97.73$\pm$0.15& \bf{97.93$\pm$0.08}& 97.42$\pm$0.63 \\
				\bottomrule
			\end{tabular}}}
	\end{minipage}}
\caption{Sensitivity analysis under varied mismatch proportions.}\label{sensitive}
\end{table}

\subsection{Visualization}\label{sec:visualization}

\begin{figure}[h]
	\centering
	\begin{minipage}{0.48\linewidth}
		\centering
		\includegraphics[width=1\linewidth]{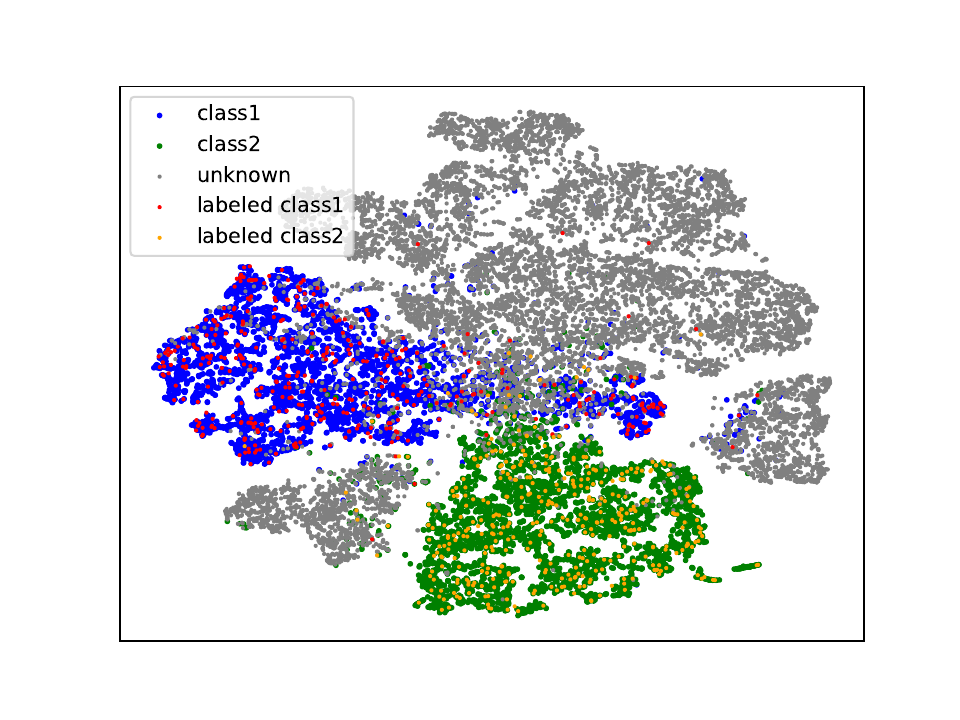}%
	\end{minipage}
	\begin{minipage}{0.48\linewidth}
		\centering
		\includegraphics[width=1\linewidth]{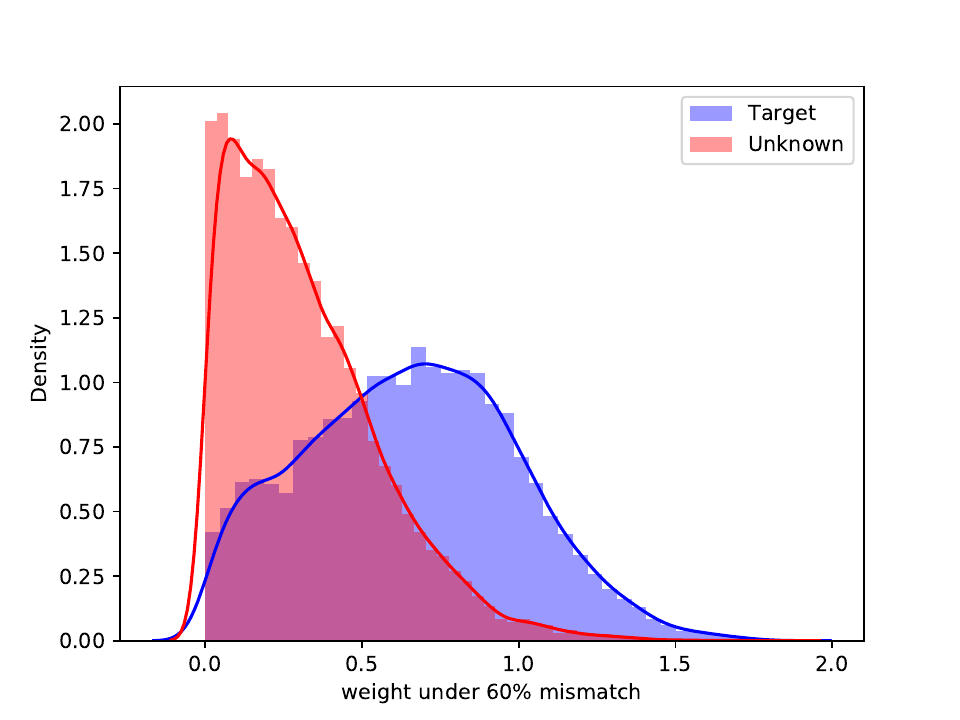}%
	\end{minipage}
	\caption{Visualization of pseudo labels and weights on CIFAR10 under 60\% class distribution mismatch.}\label{sta}
\end{figure}

\textcolor{black}{
To comprehend how WAD works, we visualize pseudo labels assigned to unlabeled instances with target categories alongside the ground truths of labeled ones in different colors, as depicted in the left part of Figure~\ref{sta}. We observe that instances with target categories are separated into two clusters and follow the same distribution as labeled instances. 
Additionally, the weight distribution, shown on the right side of Figure~\ref{sta}, depicts that WAD assigns smaller weights to unknown categories and larger ones to target ones, making it feasible to filter out harmful unknown categories and to distill useful information from target ones.
}

\section{Conclusions}


\textcolor{black}{
	To tackle class distribution mismatch in an SSL manner, we theoretically reveal that the SSL error is composed of pseudo-labeling error and invasion error under mismatch scenarios. Then, a distillation-based SSL framework, WAD, is proposed to transfer knowledge, such as pseudo labels and weights, from the representations to the target model. 
	Theoretical analyses verify that the population risk of WAD is tightly bounded. Extensive experiments on two benchmark datasets and a cross-dataset demonstrate the superiority of WAD.  
}

In the near future, we would like to investigate whether some instances from unknown categories are beneficial to target task and how to utilize them if so.

\section{Acknowledgement}

This work is supported by the National Key Research \& Develop Plan(2018YFB1004401), National Natural Science Foundation of China(62276270,62072460, 62172424), Beijing Natural
Science Foundation(4212022), Fundamental Research Funds for Central University, and the Research Funds of Renmin University of China. It is also partially supported by the Opening Fund of Hebei Key Laboratory of Machine Learning and Computational Intelligence.

{\small
\bibliographystyle{ieee_fullname}
\bibliography{egbib}

\begin{thebibliography}{10}\itemsep=-1pt

\bibitem{ba2014deep}
Jimmy Ba and Rich Caruana.
\newblock Do deep nets really need to be deep?
\newblock {\em Advances in neural information processing systems}, 27, 2014.

\bibitem{NCE_ref2}
Philip Bachman, R~Devon Hjelm, and William Buchwalter.
\newblock Learning representations by maximizing mutual information across
  views.
\newblock {\em arXiv preprint arXiv:1906.00910}, 2019.

\bibitem{berthelot2019remixmatch}
David Berthelot, Nicholas Carlini, Ekin~D Cubuk, Alex Kurakin, Kihyuk Sohn, Han
  Zhang, and Colin Raffel.
\newblock Remixmatch: Semi-supervised learning with distribution matching and
  augmentation anchoring.
\newblock In {\em International Conference on Learning Representations}, 2019.

\bibitem{berthelot2019mixmatch}
David Berthelot, Nicholas Carlini, Ian Goodfellow, Nicolas Papernot, Avital
  Oliver, and Colin~A Raffel.
\newblock Mixmatch: A holistic approach to semi-supervised learning.
\newblock {\em Advances in neural information processing systems}, 32, 2019.

\bibitem{borichev2010optimal}
Alexander Borichev and Yuri Tomilov.
\newblock Optimal polynomial decay of functions and operator semigroups.
\newblock {\em Mathematische Annalen}, 347(2):455--478, 2010.

\bibitem{food-101}
Lukas Bossard, Matthieu Guillaumin, and Luc Van~Gool.
\newblock Food-101--mining discriminative components with random forests.
\newblock In {\em European conference on computer vision}, pages 446--461.
  Springer, 2014.

\bibitem{bucilua2006model}
Cristian Buciluǎ, Rich Caruana, and Alexandru Niculescu-Mizil.
\newblock Model compression.
\newblock In {\em Proceedings of the 12th ACM SIGKDD international conference
  on Knowledge discovery and data mining}, pages 535--541, 2006.

\bibitem{ssl_cao2021open}
Kaidi Cao, Maria Brbic, and Jure Leskovec.
\newblock Open-world semi-supervised learning.
\newblock In {\em International Conference on Learning Representations}, 2021.

\bibitem{chapelle2009semi}
Olivier Chapelle, Bernhard Scholkopf, and Alexander Zien.
\newblock Semi-supervised learning (chapelle, o. et al., eds.; 2006)[book
  reviews].
\newblock {\em IEEE Transactions on Neural Networks}, 20(3):542--542, 2009.

\bibitem{30_contrastive_learning_simclr}
Ting Chen, Simon Kornblith, Mohammad Norouzi, and Geoffrey Hinton.
\newblock A simple framework for contrastive learning of visual
  representations.
\newblock In {\em International conference on machine learning}, pages
  1597--1607. PMLR, 2020.

\bibitem{28_SSL_USAD}
Yanbei Chen, Xiatian Zhu, Wei Li, and Shaogang Gong.
\newblock Semi-supervised learning under class distribution mismatch.
\newblock In {\em Proceedings of the AAAI Conference on Artificial
  Intelligence}, volume~34, pages 3569--3576, 2020.

\bibitem{Du_2022_tpami}
Pan Du, Hui Chen, Suyun Zhao, Shuwen Chai, Hong Chen, and Cuiping Li.
\newblock Contrastive active learning under class distribution mismatch.
\newblock {\em IEEE Transactions on Pattern Analysis and Machine Intelligence},
  pages 1--13, 2022.

\bibitem{Du_2021_ICCV}
Pan Du, Suyun Zhao, Hui Chen, Shuwen Chai, Hong Chen, and Cuiping Li.
\newblock Contrastive coding for active learning under class distribution
  mismatch.
\newblock In {\em Proceedings of the IEEE/CVF International Conference on
  Computer Vision (ICCV)}, pages 8927--8936, October 2021.

\bibitem{grandvalet2004semi}
Yves Grandvalet and Yoshua Bengio.
\newblock Semi-supervised learning by entropy minimization.
\newblock {\em Advances in neural information processing systems}, 17, 2004.

\bibitem{13_SSL_DS3L}
Lan-Zhe Guo, Zhen-Yu Zhang, Yuan Jiang, Yu-Feng Li, and Zhi-Hua Zhou.
\newblock Safe deep semi-supervised learning for unseen-class unlabeled data.
\newblock In {\em International Conference on Machine Learning}, pages
  3897--3906. PMLR, 2020.

\bibitem{resnet}
Kaiming He, Xiangyu Zhang, Shaoqing Ren, and Jian Sun.
\newblock Deep residual learning for image recognition.
\newblock In {\em Proceedings of the IEEE conference on computer vision and
  pattern recognition}, pages 770--778, 2016.

\bibitem{hinton2015distilling}
Geoffrey Hinton, Oriol Vinyals, and Jeff Dean.
\newblock Distilling the knowledge in a neural network.
\newblock {\em In NIPS Deep Learning and Representation Learning Workshop},
  2015.

\bibitem{infoNCE}
R~Devon Hjelm, Alex Fedorov, Samuel Lavoie-Marchildon, Karan Grewal, Phil
  Bachman, Adam Trischler, and Yoshua Bengio.
\newblock Learning deep representations by mutual information estimation and
  maximization.
\newblock In {\em International Conference on Learning Representations}, 2018.

\bibitem{ssl_huang2021trash}
Junkai Huang, Chaowei Fang, Weikai Chen, Zhenhua Chai, Xiaolin Wei, Pengxu Wei,
  Liang Lin, and Guanbin Li.
\newblock Trash to treasure: Harvesting ood data with cross-modal matching for
  open-set semi-supervised learning.
\newblock In {\em Proceedings of the IEEE/CVF International Conference on
  Computer Vision}, pages 8310--8319, 2021.

\bibitem{29_contrastive_learning_survey}
Ashish Jaiswal, Ashwin~Ramesh Babu, Mohammad~Zaki Zadeh, Debapriya Banerjee,
  and Fillia Makedon.
\newblock A survey on contrastive self-supervised learning.
\newblock {\em Technologies}, 9(1):2, 2021.

\bibitem{kim2018attention}
Wonsik Kim, Bhavya Goyal, Kunal Chawla, Jungmin Lee, and Keunjoo Kwon.
\newblock Attention-based ensemble for deep metric learning.
\newblock In {\em Proceedings of the European conference on computer vision
  (ECCV)}, pages 736--751, 2018.

\bibitem{45_adam}
Diederik~P Kingma and Jimmy Ba.
\newblock Adam: A method for stochastic optimization.
\newblock In {\em International Conference on Learning Representations}, 2015.

\bibitem{cifar10-100}
Alex Krizhevsky, Geoffrey Hinton, et~al.
\newblock Learning multiple layers of features from tiny images.
\newblock 2009.

\bibitem{laine2016temporal}
Samuli Laine and Timo Aila.
\newblock Temporal ensembling for semi-supervised learning.
\newblock {\em arXiv preprint arXiv:1610.02242}, 2016.

\bibitem{57_contrastive_review}
Phuc~H Le-Khac, Graham Healy, and Alan~F Smeaton.
\newblock Contrastive representation learning: A framework and review.
\newblock {\em IEEE Access}, 2020.

\bibitem{lee2013pseudo}
Dong-Hyun Lee et~al.
\newblock Pseudo-label: The simple and efficient semi-supervised learning
  method for deep neural networks.
\newblock In {\em Workshop on challenges in representation learning, ICML},
  volume~3, page 896, 2013.

\bibitem{miyato2018virtual}
Takeru Miyato, Shin-ichi Maeda, Masanori Koyama, and Shin Ishii.
\newblock Virtual adversarial training: a regularization method for supervised
  and semi-supervised learning.
\newblock {\em IEEE transactions on pattern analysis and machine intelligence},
  41(8):1979--1993, 2018.

\bibitem{nayak2019zero}
Gaurav~Kumar Nayak, Konda~Reddy Mopuri, Vaisakh Shaj, Venkatesh~Babu
  Radhakrishnan, and Anirban Chakraborty.
\newblock Zero-shot knowledge distillation in deep networks.
\newblock In {\em International Conference on Machine Learning}, pages
  4743--4751. PMLR, 2019.

\bibitem{flowers}
M-E Nilsback and Andrew Zisserman.
\newblock A visual vocabulary for flower classification.
\newblock In {\em 2006 IEEE Computer Society Conference on Computer Vision and
  Pattern Recognition (CVPR'06)}, volume~2, pages 1447--1454. IEEE, 2006.

\bibitem{park2019relational}
Wonpyo Park, Dongju Kim, Yan Lu, and Minsu Cho.
\newblock Relational knowledge distillation.
\newblock In {\em Proceedings of the IEEE/CVF Conference on Computer Vision and
  Pattern Recognition}, pages 3967--3976, 2019.

\bibitem{sajjadi2016regularization}
Mehdi Sajjadi, Mehran Javanmardi, and Tolga Tasdizen.
\newblock Regularization with stochastic transformations and perturbations for
  deep semi-supervised learning.
\newblock {\em Advances in neural information processing systems}, 29, 2016.

\bibitem{7diversity_coreset}
Ozan Sener and Silvio Savarese.
\newblock Active learning for convolutional neural networks: A core-set
  approach.
\newblock In {\em International Conference on Learning Representations}, 2018.

\bibitem{sohn2020fixmatch}
Kihyuk Sohn, David Berthelot, Nicholas Carlini, Zizhao Zhang, Han Zhang,
  Colin~A Raffel, Ekin~Dogus Cubuk, Alexey Kurakin, and Chun-Liang Li.
\newblock Fixmatch: Simplifying semi-supervised learning with consistency and
  confidence.
\newblock {\em Advances in neural information processing systems}, 33:596--608,
  2020.

\bibitem{tarvainen2017mean}
Antti Tarvainen and Harri Valpola.
\newblock Mean teachers are better role models: Weight-averaged consistency
  targets improve semi-supervised deep learning results.
\newblock {\em Advances in neural information processing systems}, 30, 2017.

\bibitem{wang2021knowledge}
Lin Wang and Kuk-Jin Yoon.
\newblock Knowledge distillation and student-teacher learning for visual
  intelligence: A review and new outlooks.
\newblock {\em IEEE Transactions on Pattern Analysis and Machine Intelligence},
  2021.

\bibitem{41_theoretical}
Huan Xu and Shie Mannor.
\newblock Robustness and generalization.
\newblock {\em Machine learning}, 86(3):391--423, 2012.

\bibitem{yang2022class}
Fan Yang, Kai Wu, Shuyi Zhang, Guannan Jiang, Yong Liu, Feng Zheng, Wei Zhang,
  Chengjie Wang, and Long Zeng.
\newblock Class-aware contrastive semi-supervised learning.
\newblock In {\em Proceedings of the IEEE/CVF Conference on Computer Vision and
  Pattern Recognition}, pages 14421--14430, 2022.

\bibitem{wide_resnet}
Sergey Zagoruyko and Nikos Komodakis.
\newblock Wide residual networks.
\newblock In {\em British Machine Vision Conference 2016}. British Machine
  Vision Association, 2016.

\bibitem{zhao2022decoupled}
Borui Zhao, Quan Cui, Renjie Song, Yiyu Qiu, and Jiajun Liang.
\newblock Decoupled knowledge distillation.
\newblock In {\em Proceedings of the IEEE/CVF Conference on computer vision and
  pattern recognition}, pages 11953--11962, 2022.

\bibitem{zhao2020robust}
Xujiang Zhao, Killamsetty Krishnateja, Rishabh Iyer, and Feng Chen.
\newblock Robust semi-supervised learning with out of distribution data.
\newblock {\em arXiv preprint arXiv:2010.03658}, 2020.

\bibitem{places-365}
Bolei Zhou, Agata Lapedriza, Aditya Khosla, Aude Oliva, and Antonio Torralba.
\newblock Places: A 10 million image database for scene recognition.
\newblock {\em IEEE transactions on pattern analysis and machine intelligence},
  40(6):1452--1464, 2017.

\end{thebibliography}
}

\end{document}